\documentclass{article}

\usepackage[dblblindworkshop, final]{neurips_2026}
\workshoptitle{TAE (Trust-AI-Eval): Can We Trust AI Evaluation?}

\usepackage[utf8]{inputenc}
\usepackage[T1]{fontenc}
\usepackage{lmodern}
\usepackage{hyperref}
\hypersetup{hidelinks}
\usepackage{url}
\usepackage{booktabs}
\usepackage{array}
\usepackage{amsmath,amssymb,amsfonts}
\usepackage{microtype}
\usepackage{xcolor}
\usepackage{placeins}
\usepackage{graphicx}

\title{When Pair Count Is Not the Sample Size:\\What All-Pairs Agent Comparisons Estimate}
\author{Wei-Jung Huang\\
Independent Researcher\\
\texttt{william.wj.huang@gmail.com}}

\newcommand{\E}{\mathbb{E}}
\newcommand{\Var}{\operatorname{Var}}
\newcommand{\ind}{\mathbf{1}}

\begin{document}
\maketitle

\begin{abstract}
When an agent benchmark compares every pair of leaderboard entries, the number of comparisons can look much larger than the independent evidence behind them: A versus B and A versus C both reuse A. Whether this reuse affects inference depends on what the analysis is meant to describe. If the board and its outcomes are fixed, the all-pairs mean is an exact summary of those entries, and any interval must come from another declared source of randomness. If the entries are instead treated as iid draws from a population of future configurations and the pair rule is regular and nondegenerate, the same mean is an order-two U-statistic whose first-order uncertainty depends on the number of configurations, not the number of pairs. We use near ties as the running example, but the distinction extends to other symmetric pair summaries when their regularity conditions hold. We examine both interpretations using a fixed SWE-bench Verified snapshot and an exact binary model with known truth. On SWE-bench, intervals that accounted for shared configurations were more than twice as wide as a pair-iid reference that treated the pairs as independent. In the exact model, pair-iid coverage fell far below the nominal level when edges shared endpoints but remained near nominal for matched independent edges. Results on two other fixed leaderboards show that exact summaries also depend on which pairs are included and how they are weighted. An all-pairs analysis must therefore state what is fixed, what is sampled, and how it handles shared entries and pair aggregation.
\end{abstract}

\section{Introduction}

Suppose a researcher has scored several agent configurations on the same task panel and wants to report how often two leaderboard entries are close enough to count as a tie. We use \emph{configuration} to mean one leaderboard entry. With configurations A, B, and C, the analysis contains (A,B), (A,C), and (B,C). The first two comparisons reuse A's score and task outcomes. With $M$ configurations, every entry appears in $M-1$ pairs. Counting all of those comparisons does not create an equally large set of independent configurations. All-pairs analyses already appear in model evaluation, although existing studies target different quantities and dependence structures \citep{wang2026,klabunde2025resi,ashkinaze2025deepvalue}; Section~\ref{sec:related} distinguishes those uses from the mean studied here.

Before choosing an interval, the researcher must decide what the near-tie fraction is meant to describe. When the board and task outcomes are fixed, that fraction is known exactly for the named configurations. Any interval must then reflect another declared source of randomness, such as sampled tasks, stochastic rollouts, or judges. A claim about future configurations instead treats the observed configurations as draws from a stated population. Because comparisons that share an entry can move together, the combinatorial number of pairs does not by itself justify pair-iid inference. Under iid configuration sampling and a regular nondegenerate pair rule, the all-pairs mean is an order-two U-statistic whose first-order variance is governed by the configuration count \citep{hoeffding1948,aronow2015,graham2020}. Section~\ref{sec:exact} considers independently sampled edges as a different design.

The fixed SWE-bench Verified snapshot \citep{jimenez2024} shows how accounting for shared configurations changes interval width, but it cannot reveal repeated-sampling coverage because its configurations do not come from a known population. Exact enumeration in a binary model then provides a known target for comparing shared endpoints with independent edges. Two additional snapshots from ScienceAgentBench and Terminal-Bench \citep{scienceagentbench2025,terminalbench2026} show how pair inclusion and weighting change exact fixed-board summaries. Across these settings, the results concern the interpretation of uncertainty and aggregation, not a reversal of a published leaderboard conclusion.

Our central distinction is between using the all-pairs mean to describe a named board and using it to infer properties of future configurations. Each use requires different sampling assumptions and reporting choices. Section~\ref{sec:dependence} connects this distinction to established U-statistic and dyadic variance results.

\section{Related Work}
\label{sec:related}

\paragraph{Estimands and benchmark uncertainty.}
NIST distinguishes benchmark-conditional from generalized targets and relates uncertainty to the evaluation goal \citep{keller2026}. Work on evaluation noise applies paired analysis across all model pairs, including on SWE-bench Verified, but estimates task-level noise within each pair rather than uncertainty in a mean over a sampled configuration population \citep{wang2026}. ReSi tests a correlation over all model pairs, while Deep Value Benchmark reports an all-pairs comparison that changes from $p<0.001$ to $p=0.04$ after crossed random intercepts are added for both endpoints \citep{klabunde2025resi,ashkinaze2025deepvalue}. Our analysis instead asks how the interpretation and uncertainty of an all-pairs mean change when pairwise terms reuse sampled configurations.

\paragraph{U-statistics and dyadic inference.}
Classical U-statistic theory characterizes the first-order variance of means over unordered pairs \citep{hoeffding1948}. Dyadic inference accounts for observations that share a unit, including directed and repeated dyads \citep{aronow2015,graham2020}, while crossed bootstraps resample row and column entities separately \citep{owen2007}. We use these established results to distinguish exact summaries of a named board from inference about a sampled configuration population.

\paragraph{Dependence among evaluated configurations.}
Shared model provenance can pseudo-replicate benchmark entries, and family-balanced resampling can measure sensitivity to model-family representation \citep{hardy2026,patel2026}. Because SWE lacks a validated lineage map, our empirical grouping analysis uses normalized model names only to define pair inclusion and weighting, not to support lineage-aware inference.

\section{Fixed-Board and Population Targets}
\label{sec:estimands}

The calculation has two steps: aggregate each configuration's task outcomes into a score, then compare two scores. Let $Y_{it}$ be the recorded outcome for configuration $i$ on task $t$. A fixed task aggregation produces a score $S_i$, and a symmetric rule $h(S_i,S_j)$ produces one quantity for each unordered pair. For the near-tie endpoint used below,
\begin{equation}
  h_\delta(S_i,S_j)=\ind\{|S_i-S_j|\leq\delta\}, \qquad
  U_M=\binom{M}{2}^{-1}\sum_{i<j}h_\delta(S_i,S_j).
  \label{eq:allpairs}
\end{equation}
With a fixed board and fixed outcomes, $U_M$ is the exact fraction of named configuration pairs that satisfy the rule. The pair count determines the denominator of that fraction but cannot supply uncertainty about future configurations, which requires a configuration sampling model. We use near ties because this binary rule makes both the pair construction and endpoint reuse easy to see; the same distinction between describing a fixed board and drawing inferences about a sampled population applies to other symmetric pair summaries under their corresponding regularity conditions.

The meaning of $U_M$ depends separately on whether configurations and tasks are fixed or sampled. Combining those two choices produces four cases. Table~\ref{tab:estimand-matrix} summarizes them by asking whether the claim could change with a different set of configurations or with a different task panel.\par
\begin{table}[h!]
  \centering
  \footnotesize
  \setlength{\tabcolsep}{3pt}
  \caption{The target and corresponding report for each choice of fixed or sampled configurations and tasks.}
  \label{tab:estimand-matrix}
  \begin{tabular}{>{\raggedright\arraybackslash}p{0.12\linewidth}@{\hspace{12pt}}>{\raggedright\arraybackslash}p{0.12\linewidth}>{\raggedright\arraybackslash}p{0.20\linewidth}>{\raggedright\arraybackslash}p{0.19\linewidth}>{\raggedright\arraybackslash}p{0.26\linewidth}}
    \toprule
    Entries & Tasks & Target & Randomness unit & What to report \\
    \midrule
    Fixed & Fixed & All-pairs value on the named panel & None from entries or tasks & Exact value; deletion sensitivity \\
    Fixed & Sampled & Expected pair value for the named configurations over a stated task population & Tasks or task clusters & Task-based interval; state clustering and aggregation order \\
    Sampled & Fixed & Expected pair value for a stated configuration population on the fixed tasks & Configurations or justified clusters & Shared-configuration interval; state exchangeability and regularity \\
    Sampled & Sampled & Expected pair value over stated configuration and task populations & Configurations and tasks & Joint resampling or crossed model; state aggregation order \\
    \bottomrule
  \end{tabular}
\end{table}

When configurations are fixed but tasks are sampled, uncertainty concerns a task population conditional on the named configurations. When configurations are sampled but task outcomes are fixed, the target concerns a configuration population under an explicit sampling model. If both vary, the target and procedure must represent both sources of randomness. Let $\mathcal T_R$ denote a common random panel of $R$ task clusters, and let $S_i(\mathcal T_R)$ be the resulting score for configuration $i$. For fixed configurations, one repeat-panel target is
\begin{equation}
 \theta_{\mathrm{task},R}=\E_{\mathcal T_R}\left[
 \binom{M}{2}^{-1}\sum_{i<j}h\{S_i(\mathcal T_R),S_j(\mathcal T_R)\}\right].
 \label{eq:task-estimand}
\end{equation}
When both configurations and task panels are sampled, the analogous joint target is
\begin{equation}
 \theta_{\mathrm{joint},R}=\E_{X_1,X_2,\mathcal T_R}
 [h\{S(X_1,\mathcal T_R),S(X_2,\mathcal T_R)\}].
\end{equation}
Both targets first aggregate outcomes on a common task panel and then apply the pair kernel. When $h$ is nonlinear, changing that order can give a different answer, whether by applying $h$ to task-population mean scores or by averaging a task-level pair quantity. The estimand therefore includes the panel size, cluster design, sharing of the common panel, and aggregation order. These four sampling regimes need not produce the same interval, and enumerating every edge does not select one of them. Evaluation guidance similarly ties uncertainty to both the estimand and the benchmark entities treated as sampled \citep{keller2026}.

\FloatBarrier

\section{Dependence from Shared Configurations}
\label{sec:dependence}

If configuration A is unusually likely to form near ties, comparisons involving A tend to move together because they all reuse the same entry. The following decomposition quantifies that shared contribution. For inference about future configurations, let $X_1,\ldots,X_M$ be independent and identically distributed configuration records, let $\theta=\E[h(X_1,X_2)]$, and define the first-order projection
\begin{equation}
  g(x)=\E[h(x,X_2)]-\theta, \qquad \zeta_1=\Var(g(X_1)).
\end{equation}
The Hoeffding decomposition gives
\begin{equation}
  U_M-\theta=\frac{2}{M}\sum_{i=1}^{M}g(X_i)+R_M,
  \qquad
  \Var(U_M)=\frac{4\zeta_1}{M}+O(M^{-2}),
  \label{eq:rate}
\end{equation}
when the kernel is regular, meaning $\zeta_1>0$, and standard moment conditions hold. Each first-order term appears in $M-1$ pairs, which gives the coefficient $2/M$, while the degenerate residual contributes $O(M^{-2})$ variance. Under the corresponding conditions, Hoeffding's classical result gives the $M^{-1}$ variance order and asymptotic normality \citep{hoeffding1948}. A pair-iid calculation instead treats the $N=\binom{M}{2}$ pair values as independent, giving variance of order $N^{-1}=O(M^{-2})$. These rates differ because $O(M^3)$ pairs of edges share an endpoint. Dyadic sandwich estimators account for correlation among dyads that share a member \citep{aronow2015,graham2020}.

Equation~\eqref{eq:rate} applies to future-configuration claims only after the population, exchangeable unit, and regular first-order regime have been specified. If configurations share model families or contributors, one configuration may not be the appropriate exchangeable unit \citep{hardy2026,patel2026}. Randomness from tasks, rollouts, and judges introduces separate dimensions. Crossed-array procedures can resample row and column entities separately \citep{owen2007}, but only when those entities represent the populations named by the estimand.

The sample diagnostic below checks whether the first-order variance estimate is positive and accounts for at least half of the total estimated variance. These sample conditions indicate whether the regular first-order formulas are usable for the observed sample; exchangeability and population regularity still depend on the design. Section~\ref{sec:exact} therefore separates repeated-sampling coverage from diagnostic behavior, and Appendix~\ref{app:details} gives the formulas for each method.

\section{Interval Width and Coverage}
\label{sec:swe}

\subsection{SWE-bench Verified: Interval Width}

The SWE-bench Verified panel contains 134 configurations evaluated on the same 500 tasks, yielding 67,000 binary configuration-task cells without duplicates. We set the near-tie margin in Eq.~\eqref{eq:allpairs} to $\delta=0.05$, for which the point estimate is an exact description of this panel. To examine interval width under a future-configuration interpretation, we adopt a working model that treats the 134 configurations as independent draws and compare four intervals that account for their reuse across pairs with a pair-iid Wald reference. Because the board provides no sampling design for future submissions, we treat this comparison as a sensitivity analysis under the working model rather than as a design-based population analysis. Separately, the width curves mark 0.05 as an illustrative reference value, not a general standard for sufficient precision.

\begin{table}[ht]
\centering
\small
\caption{Near-tie estimate and intervals for the fixed SWE-bench Verified panel of 134 configurations. Pair-iid uses 8,911 unordered edges; other rows account for shared configurations under the working model. Width ratios use unrounded widths relative to pair-iid.}
\label{tab:swe-sensitivity}
\begin{tabular}{lrrrr}
\toprule
Method & Estimate & 95\% interval & Width & Ratio $\times$ \\
\midrule
Pair-iid reference & 0.159 & [0.152, 0.167] & 0.015 & 1.000 \\
Hoeffding projection & 0.159 & [0.141, 0.177] & 0.036 & 2.354 \\
Dyadic sandwich & 0.159 & [0.140, 0.178] & 0.038 & 2.511 \\
Delete-one-configuration jackknife & 0.159 & [0.138, 0.180] & 0.042 & 2.733 \\
Off-diagonal node bootstrap & 0.159 & [0.139, 0.182] & 0.043 & 2.860 \\
\bottomrule
\end{tabular}
\end{table}

Accounting for shared configurations more than doubles the interval width. The finite-board near-tie rate is 0.159, for which the pair-iid reference has total width 0.0152. Under the working model of independent configurations, the four shared-configuration widths range from 0.0358 to 0.0435, or 2.35 to 2.86 times the reference width (Table~\ref{tab:swe-sensitivity}). All five widths fall below the illustrative 0.05 reference at the chosen margin.

\begin{figure}[t]
  \centering
\includegraphics[width=\linewidth]{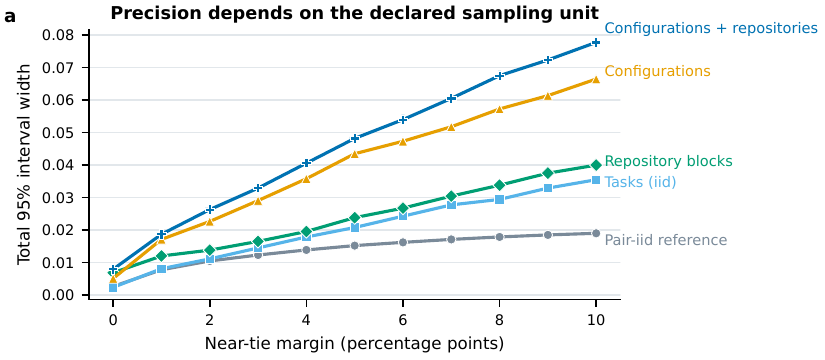}
  \caption{Total widths of exploratory 95 percent sensitivity intervals for the fixed SWE-bench Verified panel across near-tie margins. Pair-iid is an unclipped Wald reference over 8,911 edges; the other curves use equal-tail type-7 percentile bootstraps over tasks, 12 repository blocks, configurations, or configurations and repositories jointly. Each bootstrap uses 2,000 replicates on 134 configurations and 500 tasks. The horizontal line marks width 0.05.}
  \label{fig:estimand-sensitivity}
\end{figure}

Varying the near-tie margin shows where the interval widths cross the illustrative reference. Joint resampling first crosses 0.05 at a six-percentage-point margin, configuration resampling crosses it at seven points, and the pair-iid reference remains below it through 10 points. Figure~\ref{fig:estimand-sensitivity} displays the complete width curves. At the five-percentage-point margin, resampling tasks or repository blocks gives widths of 0.0208 and 0.0238; resampling configurations gives 0.0435, and jointly resampling configurations and repository blocks gives 0.0482. The pair-iid reference is narrowest at all 10 positive margins, although resampling individual tasks gives a slightly narrower interval at zero. Because the curves represent different sampled populations, their widths do not rank interchangeable methods. Lineage resampling with singleton labels is identical to configuration resampling, so the figure omits the duplicate curve.

Task thinning and leave-one-configuration-out deletion quantify how the fixed estimate changes when half the tasks or one entry is removed. Conditional on the observed outcomes, each of 2,000 Monte Carlo replicates sampled 250 of the 500 tasks without replacement and used the same subset for every pair. We call an ordering a strict reversal when the subset ordering is opposite to the ordering on all tasks; an exact tie in either panel remains unresolved. The mean strict-reversal and unresolved rates were 0.0189 and 0.0101, and removing any one configuration changed the near-tie estimate by at most 0.00230.

The available metadata are not sufficient for a lineage analysis. Among the 134 configurations, 43 lacked a model tag, while a proxy based on submission site or organization labeled 122 configurations but placed 29 in one group. These fields do not establish genealogy, so a lineage-sensitive summary would require validated family labels and is not reported.

\FloatBarrier

\subsection{Exact Coverage Under Shared and Independent Edges}
\label{sec:exact}

The fixed SWE panel shows how reused configurations change interval width, but its configurations do not come from a known population and therefore cannot reveal repeated-sampling coverage. To study coverage with known truth, we use a binary model and compare two constructions that differ only in whether pair outcomes share endpoints. In the regular equality model, $X_i\sim\mathrm{Bernoulli}(1/5)$ independently, and a pair is marked when the two binary values match, $H_{ij}=\ind\{X_i=X_j\}$, giving target $\theta=17/25$. The iid generator and positive population first-order variance justify regular first-order analysis in this model. We treat the enumeration as retrospective and exploratory because the grid was selected after prior results were available. Given $n=\sum_i X_i$,
\begin{equation}
 U_M=\frac{\binom{n}{2}+\binom{M-n}{2}}{\binom{M}{2}}.
 \label{eq:equality}
\end{equation}
For $M\in\{40,80,160\}$, we enumerate every $n\in\{0,\ldots,M\}$ with its exact binomial probability and compare a pair-iid Wald interval with the Hoeffding projection, dyadic sandwich, and delete-one-unit jackknife intervals. Unconditional coverage is the primary repeated-sampling quantity. We also record how often realized samples pass the variance diagnostic and the coverage among those samples; Appendix~\ref{app:details} defines both secondary quantities.

\paragraph{Regular equality model.}
As $M$ grows from 40 to 160, the apparent pair count increases from 780 to 12,720, yet pair-iid coverage falls from 0.307 to 0.157 (Table~\ref{tab:exact-characterization}). The shared-endpoint methods attain unconditional coverage from 0.928 to 0.940, which is closer to the nominal level but remains below it on this grid. Although their variance estimates differ, the three methods have identical covering sets here; this agreement is limited to the enumerated grid.

\paragraph{Matched independent-edge control.}
The matched control keeps the target $17/25$, nominal pair counts, values of $M$, critical value, and pair-iid interval formula fixed, but replaces the shared-endpoint outcomes with $N=\binom{M}{2}$ independent Bernoulli edges. Coverage is 0.950 to 0.951 across the three values of $M$, and its gap from the shared-endpoint pair-iid coverage grows from 0.642 to 0.794. Wald discreteness, the binary outcome, the target, and the pair count are unchanged. Within this model, the coverage gap is therefore attributable to whether the edges reuse endpoints.

\begin{table}[ht]
\centering
\footnotesize
\setlength{\tabcolsep}{2pt}
\caption{Exact coverage for the shared-endpoint grid and matched independent-edge control. Coverage probabilities come from full enumeration rather than empirical confidence intervals and are rounded to three decimals.}
\label{tab:exact-characterization}
\begin{tabular}{lrrrrrr}
\toprule
Scenario & $M$ & Pairs & \shortstack{Shared edges,\\pair-iid} & \shortstack{Shared edges,\\endpoint-aware} & \shortstack{Independent edges,\\pair-iid} & Gap \\
\midrule
Equality & 40 & 780 & 0.307 & 0.928 & 0.950 & 0.642 \\
Equality & 80 & 3160 & 0.220 & 0.932 & 0.950 & 0.730 \\
Equality & 160 & 12720 & 0.157 & 0.940 & 0.951 & 0.794 \\
\bottomrule
\end{tabular}
\par\smallskip\footnotesize The first two coverage columns use shared edges with pair-iid and endpoint-aware formulas. The third uses independent Bernoulli edges with the same target, pair count, critical value, and pair-iid formula. Gaps use unrounded coverage.
\end{table}

\FloatBarrier

As a boundary case, when $H_{ij}=1$ for every pair, all evaluated intervals equal $[1,1]$ and cover unconditionally, but zero first-order variance prevents the three shared-configuration methods from passing the diagnostic, so this result does not establish regular calibration.

\FloatBarrier

\section{Sensitivity to Pair Inclusion and Weighting}
\label{sec:block-aggregation}

Even an exact fixed-board description depends on which pairs are included and how they are weighted. If one normalized model name appears under many configurations while another appears once, the repeated name contributes more terms to an all-configuration-pair mean. We compare that mean, $A$, with an alternative, $E$, that excludes within-block pairs and gives every pair of distinct model-name blocks equal weight. The difference therefore combines pair exclusion with reweighting; it is not a lineage effect. Appendix~\ref{app:details} gives the formal definition.

Our two primary snapshots come from ScienceAgentBench and Terminal-Bench \citep{scienceagentbench2025,terminalbench2026}, each restricted to common task support. Table~\ref{tab:fixed-panel-weighting} also includes panels from SciCode, AssistantBench, and CORE-Bench \citep{tian2024scicode,yoran2024assistantbench,siegel2024corebench}, along with nested ScienceAgentBench strata. The three additional panels serve as two small stress panels and a unique-block control; they clarify how the aggregation behaves but are not independent replications.

\begin{table}[h!]
\centering
\footnotesize
\setlength{\tabcolsep}{2.5pt}
\caption{Exact near-tie rates at the five-percentage-point margin under all-configuration and equal-block aggregation for fixed public snapshots and stated strata.}
\label{tab:fixed-panel-weighting}
\begin{tabular}{lrrrrrr}
\toprule
Panel & Configs & Blocks & Pairs & All-pair (\%) & Equal-block (\%) & $A-E$ (pp) \\
\midrule
ScienceAgentBench & 23 & 11 & 253 & 33.60 & 30.68 & 2.92 \\
Self-Debug subset & 16 & 11 & 120 & 50.00 & 40.91 & 9.09 \\
Generalist subset & 7 & 5 & 21 & 28.57 & 25.00 & 3.57 \\
SciCode (stress) & 10 & 7 & 45 & 75.56 & 81.75 & -6.19 \\
AssistantBench (stress) & 9 & 5 & 36 & 69.44 & 52.22 & 17.22 \\
CORE-Bench Hard (control) & 7 & 7 & 21 & 9.52 & 9.52 & 0.00 \\
Terminal-Bench S43 & 43 & 22 & 903 & 16.72 & 11.95 & 4.78 \\
\bottomrule
\end{tabular}
\par\medskip
\begin{minipage}{\linewidth}
\footnotesize
Counts are exact; rates and percentage-point differences are rounded to two decimals.
\end{minipage}
\end{table}

On the two primary snapshots, $A-E$ is 2.92 percentage points for ScienceAgentBench and 4.78 percentage points for Terminal-Bench. The nested and stress rows vary in sign because excluding within-block pairs and reweighting block pairs can raise or lower the mean depending on where near ties occur. The unique-block CORE-Bench Hard control has $A=E$ exactly. These values are exact descriptions of named snapshots, so no sampling interval is attached.

\FloatBarrier

\section{Reporting Implications for Agent Evaluation}
\label{sec:implications}

An all-pairs mean does not determine its own uncertainty: on a named board, the value is exact once the task aggregation, pair rule, included pairs, and weights are fixed, whereas under a population interpretation, uncertainty comes from whichever configurations, tasks, rollouts, or judges are treated as sampled. The combinatorial pair count describes how many terms enter the mean, not which observations provide independent evidence.

These results imply four reporting choices:
\begin{enumerate}\setlength{\itemsep}{0pt}\setlength{\parsep}{0pt}\setlength{\topsep}{2pt}
  \item For a fixed-board target, name the configurations and tasks, specify the order of task aggregation and pair comparison, and state which pairs are included, how they are weighted, and how ties and missing outcomes are handled.
  \item For a population target, name each sampling source, the population it represents, and the exchangeable unit, including how shared families, contributors, or deployments could violate exchangeability.
  \item Match the variance formula or resampling scheme to the sampled entities, and report the assumptions and diagnostic used to justify a regular interval alongside coverage or width. The constant-kernel boundary shows why neither unconditional coverage nor zero width alone establishes regular first-order calibration. No interval family is valid for every all-pairs analysis because the appropriate interval follows from the target and sampling design.
  \item When a simulation or exact model permits a matched independent-edge control, report whether the discrepancy remains after replacing shared endpoints with independently sampled edges. Holding the target, pair count, critical value, and interval formula fixed isolates endpoint reuse within that design.
\end{enumerate}
\section{Limitations}
\label{sec:limitations}

The benchmark results support exact descriptions of the named panels and sensitivity analyses under the stated working models. Without a sampling frame or defensible exchangeability assumptions, they cannot support population claims about future submissions. For SWE, incomplete lineage, missing repeated outcomes, and unavailable judge records mean that a configuration may not be the appropriate resampling unit. ScienceAgentBench and Terminal-Bench rely on retrospective name grouping without nested replication, and Terminal-Bench also mixes rollout depths. Differences in metrics, configuration support, and observable randomness across the analyzed benchmarks also preclude a pooled cross-benchmark estimate.

The exact coverage results apply only to the enumerated binary models, values of $M$, and interval formulas. Because the equality and independent-edge grids were selected after outcomes were available, this part of the analysis is retrospective. The constant kernel marks the point at which regular first-order reasoning no longer applies; studying nonregular methods would require a different analysis. Likewise, the matched control compares complete endpoint reuse with independent edges but does not trace intermediate levels of dependence. That question would require a prespecified family of joint edge distributions rather than another post hoc condition.

This paper does not establish how often published agent evaluations use pair-iid inference or whether correcting it would overturn conclusions. Answering either question would require a systematic review of inferential units and interval calculations.

\section{Conclusion}

An all-pairs mean describes different targets depending on what the analysis treats as sampled: on a fixed board, it exactly describes the named entries under the chosen aggregation, whereas under iid configuration sampling and a regular nondegenerate kernel, its first-order variance scales as $M^{-1}$ rather than with the combinatorial pair count. Exact enumeration shows that pair-iid coverage deteriorates when edges share endpoints but remains near the nominal level for matched independent edges. On the real panels, accounting for shared configurations widened intervals, and changing pair inclusion and weighting altered exact summaries. These results support fixed-board descriptions and sensitivity analyses under the stated working models, not population claims about future submissions without a sampling frame or defensible exchangeability assumptions; an all-pairs report should therefore identify its target, sampled entities, pair weights, and matching uncertainty design.

\bibliographystyle{plainnat}
\bibliography{references}

\clearpage
\appendix
\section{Additional Methods and Results}
\label{app:details}

\subsection{Fixed-Panel Construction}

The empirical matrix contains every configuration with a complete outcome vector on the common 500-task SWE-bench Verified snapshot. Task and immutable submission identifiers were normalized with Unicode NFKC, with outer whitespace removed and case preserved. A configuration was eligible only if its normalized submission identifier was unique and appeared in both the outcome and metadata records, every one of the 500 task identifiers had exactly one binary result, and no additional task identifier appeared. An identifier collision, duplicate configuration-task cell, or metadata mismatch made a configuration ineligible. All 134 configurations passed, yielding 67,000 binary configuration-task cells and 8,911 unordered configuration pairs. The inclusion rule did not use endpoint values or leaderboard ranks. We define a configuration's primary score as its mean binary outcome across the 500 fixed tasks. If $c_i$ is the success count for configuration $i$, the exact integer rule $20|c_i-c_j|\leq500$ classifies a pair as a near tie. This is equivalent to requiring an absolute score difference of at most 0.05. The point estimate was unchanged when eligible pairs were duplicated as directed comparisons or their row order was permuted.

The fixed-board quantity is
\begin{equation}
 \widehat{\theta}_{\mathrm{board}}=
 \frac{1}{\binom{134}{2}}\sum_{i<j}
 \ind\{|S_i-S_j|\leq 0.05\}.
\end{equation}
We calculate this quantity exactly, without stochastic approximation, and use intervals over configurations only as sensitivities under a working model. The analytic intervals display the two-sided normal critical value as 1.96 but use its full machine-precision value in the calculation; the off-diagonal configuration bootstrap uses percentile endpoints.

\subsection{Interval and Influence Calculations}

Let $H_{ij}=H_{ji}=h(S_i,S_j)$, $N=\binom{M}{2}$, and $\bar H=U_M$. The pair-iid reference treats the $N$ kernel values as independent and uses
\begin{equation}
 \widehat V_{\mathrm{pair}}=
 \frac{1}{N(N-1)}\sum_{i<j}(H_{ij}-\bar H)^2.
 \label{eq:pair-iid-variance}
\end{equation}
All normal intervals use $\bar H\pm z\sqrt{\widehat V}$ without clipping, where $z$ is the full-precision two-sided 95 percent normal critical value.

For the Hoeffding projection, let $\mathcal E$ be the set of unordered configuration pairs and define
\begin{align}
 \widehat A &={\left\{M\binom{M-1}{2}\right\}^{-1}}
 \sum_{i=1}^{M}\sum_{\substack{j<k\\j,k\ne i}}H_{ij}H_{ik}, \\
 \widehat B &={\left\{3\binom{M}{4}\right\}^{-1}}
 \sum_{\substack{\{e,e'\}\subset\mathcal E\\e\cap e'=\varnothing}}H_eH_{e'},
 \qquad
 \widehat C=N^{-1}\sum_{i<j}H_{ij}^2. 
\end{align}
We compute the numerator of $\widehat B$ as the sum over all unordered pairs of edges minus the adjacent-edge contribution, avoiding explicit enumeration of the $3\binom{M}{4}$ disjoint edge pairs. We set $\widehat\zeta_{1,+}=\max(0,\widehat A-\widehat B)$ and $\widehat\zeta_{2,+}=\max(0,\widehat C-2\widehat A+\widehat B)$, then use
\begin{equation}
 \widehat V_{\mathrm{H}}=\frac{4\widehat\zeta_{1,+}}{M}.
 \label{eq:hoeffding-variance}
\end{equation}

For the intercept-only dyadic sandwich, write $r_{ij}=H_{ij}-\bar H$, $d_i=\sum_{j\ne i}r_{ij}$, and $Q=\sum_{i<j}r_{ij}^2$. Its variance estimate is
\begin{equation}
 \widehat V_{\mathrm{D}}=
 \max\left\{0,\frac{\sum_i d_i^2-Q}{N^2}\right\}.
 \label{eq:dyadic-variance}
\end{equation}
For the delete-one-configuration jackknife, let $U_{(-i)}$ be the all-pairs mean after removing configuration $i$, $P_i=M\bar H-(M-1)U_{(-i)}$, and $\bar P=M^{-1}\sum_iP_i$. We use
\begin{equation}
 \widehat V_{\mathrm{J}}=
 \frac{1}{M(M-1)}\sum_{i=1}^{M}(P_i-\bar P)^2.
 \label{eq:jackknife-variance}
\end{equation}
For the off-diagonal bootstrap, each draw samples multinomial counts $(w_1,\ldots,w_M)$ from $M$ draws with equal probabilities and computes
\begin{equation}
 U_M^*=\frac{\sum_{i<j}w_iw_jH_{ij}}{\sum_{i<j}w_iw_j}.
 \label{eq:node-bootstrap}
\end{equation}
We redraw samples with a zero denominator and take the type-7 sample quantiles at 0.025 and 0.975 as the interval endpoints. Among the 2,000 bootstrap draws, 1,810 values were distinct.

Let $\widehat\zeta_{1,+}$ and $\widehat\zeta_{2,+}$ denote the nonnegative estimates of the first-order and second-order variance components. The estimated first-order share was
\begin{equation}
 D=\frac{4\widehat\zeta_{1,+}/M}
 {4\widehat\zeta_{1,+}/M+2\widehat\zeta_{2,+}/\{M(M-1)\}}=0.852.
\end{equation}
The panel-level condition $D\geq 0.5$ was satisfied. The relevant raw variance or jackknife standard error was positive for each analytic method, and the bootstrap produced at least two distinct values. We therefore report all four methods as regular working-model sensitivities under the stated independent-configuration model. The observed leaderboard cannot establish independent sampling or exchangeability, and $D<0.5$ by itself would not prove population degeneracy. Deleting one configuration changed the estimate by at most 0.00230 in absolute value.

We also computed task-only bootstrap intervals by resampling either individual tasks or the 12 repository blocks. Because tasks are clustered within repositories, the individual-task bootstrap width of 0.0208 serves only as a comparator. The repository-block bootstrap preserved each repository's observed task count and had width 0.0238. Independently resampling configurations and repository blocks gave a joint sensitivity width of 0.0482. Each calculation therefore corresponds to a different source of randomness.

\subsection{Half-Task Sensitivity}

In each of 2,000 Monte Carlo replicates, we sampled 250 of the 500 tasks uniformly without replacement and used the same subset for every unordered configuration pair. We compared each pair's ordering on the subset with its ordering on all 500 tasks, treating an exact tie in either panel as unresolved. Conditional on the fixed outcome panel, the mean strict-reversal and unresolved rates were 0.0189 and 0.0101. The central 95 percent type-7 sample quantiles of the strict-reversal rates were $[0.0149,0.0237]$. This range measures sensitivity to random half-task thinning, not uncertainty in a latent parameter or Monte Carlo error in the reported mean.

Pairing every half-task draw with independent configuration weights gave the joint sensitivity interval $[0.0123,0.0276]$. This sensitivity analysis under the working model represents sampling both configurations and tasks and is reported separately from the illustrative width comparison in the main text.

\subsection{Available Lineage Metadata}

The available metadata do not support a reliable lineage grouping. Exact normalized model tags covered 91 of the 134 configurations, leaving 43 unresolved across 55 observed tags. A second proxy used the submission-site hostname, with organization as a fallback; it labeled 122 configurations but placed 29 in one group. Neither proxy is a validated genealogy of base models and agents, so we did not analyze lineage-sensitive outcomes or report a lineage-aware interval.

\subsection{Alternative Pair Inclusion and Weighting}

We apply the alternative pair-inclusion and weighting rule retrospectively to the same inspected records. Let $G_1,\ldots,G_B$ be normalized model-name blocks with sizes $n_1,\ldots,n_B$. Alongside the all-configuration-pair mean $A=U_M$, we compute
\begin{equation}
 E=\binom{B}{2}^{-1}\sum_{b<c}\frac{1}{n_bn_c}
 \sum_{i\in G_b}\sum_{j\in G_c}h(S_i,S_j).
 \label{eq:equal-block}
\end{equation}
The quantity $E$ excludes within-block pairs, averages the comparisons within each pair of distinct blocks, and then gives every block pair equal weight. These blocks use normalized names as descriptive metadata, not as lineages.

The ScienceAgentBench live-page panel contains 23 configurations with complete binary outcomes on 102 common tasks and 11 normalized model-name blocks. Its 253 unordered configuration pairs give $A=85/253$; the 55 distinct block pairs give $E=27/88$. The two displayed ScienceAgentBench strata partition the 23 configurations into groups of 16 and 7 on the same task panel. The outcomes are historical HAL evaluations that predate the April 2026 verified benchmark update.

The primary Terminal-Bench S43 panel contains 43 configurations on 89 common task keys, 22 normalized model-name blocks, and 903 unordered configuration pairs. It includes 15 configurations with one rollout per task, one with four, and 27 with five. We averaged outcomes within each task before giving every task equal weight. The exact fixed-set values are $A=151/903$ and $E=1987/16632$. Retained records connect the pinned leaderboard revision to the configuration metadata, selected jobs, aggregate results, and reduced trial records used for this panel. Because the historical task payload bytes were not retained, these records cannot establish byte identity for the benchmark task definitions.

A separate Terminal-Bench sensitivity gave a pair-iid width of 0.0487 and a delete-one-configuration width of 0.102. Both calculations use the same retrospective snapshot and define no interval for future configurations.

The table also reports two small stress panels, SciCode and AssistantBench, and one unique-block control, CORE-Bench Hard. The control has seven configurations in seven blocks and yields $A=E=2/21$. Differences in the nested and stress panels vary in direction because the alternative can raise or lower the mean depending on where near ties occur. All rows describe observed fixed sets and therefore carry no sampling interval.

\subsection{Exact Enumeration and Diagnostics}

For the equality kernel, the sufficient state is $n=\sum_i X_i$ with exact probability
\begin{equation}
 \Pr(n)=\binom{M}{n}\frac{4^{M-n}}{5^M}.
\end{equation}
The enumeration includes every state $n=0,\ldots,M$. For each interval, we determine coverage using an integer form of
\begin{equation}
 (U_M-\theta)^2\leq z^2\widehat V,
\end{equation}
with equality counted as coverage. When $\widehat V=0$, an interval covers only if $U_M=\theta$ exactly. Evaluating four methods at $M=40,80,160$ gives 2,264 method-state rows across the two kernels. This characterization uses neither random numbers nor Monte Carlo estimates.

Let $E_m$ denote a pass of the diagnostic and $C_m$ denote coverage by method $m$'s interval. The diagnostic uses the variance estimates defined above. We report four distinct quantities:
\begin{equation}
 \Pr(C_m), \qquad \Pr(E_m), \qquad
 \Pr(C_m\cap E_m), \qquad
 \Pr(C_m\mid E_m)=\frac{\Pr(C_m\cap E_m)}{\Pr(E_m)}.
 \label{eq:eligibility}
\end{equation}
Coverage among passing states is undefined when $\Pr(E_m)=0$. Unconditional coverage remains the primary repeated-sampling property. The other three quantities record how often samples meet the diagnostic and how coverage behaves among those samples; none verifies the design assumption of exchangeability.

The sample diagnostic requires a positive first-order variance estimate and a first-order share of at least one half. For the equality kernel, the probability of passing rounds to at least 0.999 at all three values of $M$. Regular first-order analysis is justified by the iid Bernoulli model and its positive population first-order variance, independently of whether a particular sample passes the diagnostic. Under the constant kernel, every variance estimate and interval width is zero. At $M=40,80,160$, corresponding to 780, 3,160, and 12,720 pairs, both pair-iid and endpoint-aware coverage equal 1.000. The three shared-configuration methods have zero probability of passing the diagnostic, while pair-iid remains a defined reference calculation; no matched independent-edge control is defined for this boundary. All of these quantities refer only to the enumerated model.

\subsection{Conditional Coverage and Diagnostics}

The exact grid was designed after prior simulation outcomes were available, making this analysis retrospective and exploratory; we therefore report every evaluated cell and method. Among passing samples, conditional coverage for the Hoeffding projection, dyadic sandwich, and jackknife was 0.929 at $M=40$ and lay in $[0.925,0.975]$ at $M=160$, while the pair-iid mismatch persisted. Under the constant kernel, unconditional coverage remained one and the probability of passing the diagnostic remained zero. These results characterize only the evaluated grid.

\subsection{Exact Comparison with Independent Edges}

The matched control is exploratory because it was specified after the shared-endpoint results were known. For each $M\in\{40,80,160\}$, let $N=\binom{M}{2}$ and draw $N$ independent edge outcomes with success probability $\theta=17/25$. If $B$ denotes their success count, then $U=B/N$ and the pair-iid interval is
\begin{equation}
 U\mathbin{\pm}1.96
 \sqrt{\frac{U(1-U)}{N-1}},
\end{equation}
without endpoint clipping. We display the critical value as 1.96 but use its full machine-precision value in the calculation. Every $B=0,\ldots,N$ enters the enumeration with its exact integer Binomial mass. For $M=40,80,160$, the covering-state sets contained 51, 103, and 207 counts. Exact coverage was 0.950, 0.950, and 0.951, and the corresponding expected widths were 0.0655, 0.0325, and 0.0162. High-precision decimal and binary64 calculations classified every state identically; the integer probability masses summed exactly to $25^N$, and a six-edge brute-force enumeration matched the sufficient-count calculation.

\subsection{Numerical Validation}

We compared every exact state with the corresponding binary64 calculation and a rational oracle. Coverage classifications and diagnostic decisions had zero disagreements, and the largest absolute binary64 error was $7.77\times 10^{-15}$. A separate control used $|13/20-12/20|\leq0.05$, which is true in exact arithmetic. Binary64 represented the difference as $0.050000000000000044$ and returned false for a direct comparison, whereas the integer rule used in the analysis returned true. This boundary case checks the implementation rather than the statistical mechanism.

All analyses ran on a Windows x64 machine with eight logical Intel CPU processors, no accelerator or model calls, and an 8 GB memory cap. Total runtime was 256 wall-clock seconds and 237 CPU seconds; peak working-set memory was unavailable. The exact shared-endpoint and matched independent-edge enumerations took 1.59 and 1.56 seconds, respectively, under a 2 GB memory cap.

\end{document}